\documentclass{article}

\usepackage[preprint]{neurips_2020}

\usepackage{natbib}
\setcitestyle{numbers}

\usepackage[utf8]{inputenc} 
\usepackage[T1]{fontenc}    
\usepackage{url}            
\usepackage{booktabs}       
\usepackage{amsfonts}       
\usepackage{nicefrac}       
\usepackage{graphicx}
\usepackage{enumitem}
\usepackage{wrapfig,lipsum}
\usepackage{textcomp}
\usepackage{multirow}
\usepackage{caption}
\usepackage{subcaption}
\usepackage{makecell}
\usepackage{booktabs}
\usepackage{hyperref}       

\title{Building Goal-Oriented Dialogue Systems with Situated Visual Context}

\author{%
  Sanchit Agarwal \And Jan Jezabek \And Arijit Biswas \And Emre Barut \And Shuyang Gao \And Tagyoung Chung \\ \\
  Amazon Alexa AI \\ \\
  \texttt{\{agsanchi, jezabek, barijit, ebarut, shuyag, tagyoung\}}@amazon.com
}

\begin{document}

\maketitle

\begin{abstract}
Most popular goal-oriented dialogue agents 
are
capable of understanding the conversational context. However, with the surge of
virtual assistants with screen, the next generation of agents are required to also
understand screen context in order to provide a proper interactive experience, and better understand users' goals. In this paper, we propose a novel multimodal
conversational framework, where the dialogue agent's next action and their arguments are
derived jointly conditioned both on the conversational and the visual
context. Specifically, we propose a new model, that can reason over the visual
context within a conversation and populate API arguments with visual entities
given the user query. Our model can recognize visual features such as color
and shape as well as the metadata based features such as price or star rating
associated with a visual entity. In order to train our model, due to a lack of suitable multimodal
conversational datasets, we also propose a novel multimodal dialog
simulator to generate synthetic data and also collect realistic user data from MTurk to improve model robustness. The proposed model achieves a reasonable 
model accuracy, without high inference latency. We also demonstrate the proposed approach in a prototypical furniture shopping experience for a multimodal virtual assistant.
\end{abstract}

\section{Introduction}
Goal-oriented dialogue systems enable users to complete specific goals such as
booking a flight. The user informs their intent and the agent
will ask for the slot values such as time and number of people before booking. The traditional goal-oriented systems are usually aware of the
conversational context between the chatbot and the
user~\cite{acharya-etal-2021-alexa, Wen2017, liu2017end, Shah2018}, where users
can refer to entities from the context. However, with the recent surge of
virtual assistants with screens (e.g., Echo Show), the screen becomes an additional source of context that is vital in fulfilling the users' goal. For example, when a user browsing for chairs on Echo Show says, \emph{``What is the price of the black checkered one?''}, the virtual assistant needs to identify the right product based on its visual characteristics and then respond with the price. Moreover, users may want to visually compare items from the current screen with those shown on earlier screens by asking questions such as \emph{``Is this cheaper than the green t-shirt you showed earlier?''}. The dialogue agent needs to understand the multimodal context in order to resolve the relevant entities and then compare their prices. Since, most current dialogue systems do not take the visual context into account, they are unable to fulfill such goals.

\begin{figure}
\centering
\includegraphics[width=0.55\linewidth]{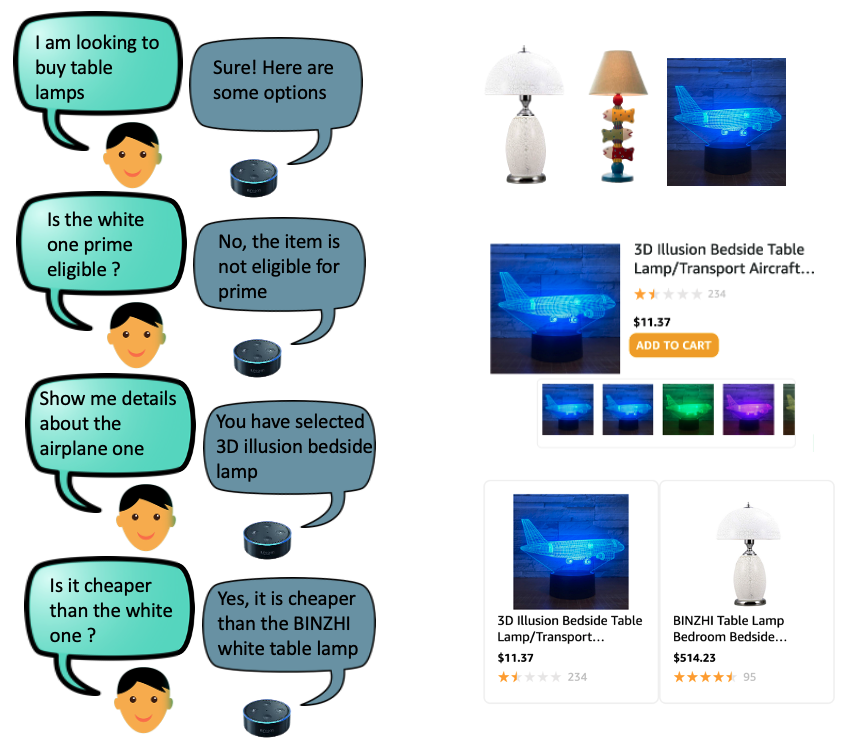}
\caption{\small Example Multimodal Conversation.}\label{fig:example_multimodal_conversation}
\end{figure}

In this paper, we propose a novel multimodal goal-oriented dialogue system that
can reason over the current and historical screen context as well as the
conversational context in order to complete users' goal. An example conversation
with an agent built using our proposed method is demonstrated in
Figure~\ref{fig:example_multimodal_conversation}. Users can refer to on-screen
visual entities by attributes such as color (e.g., ``the white one'') or shape (e.g.,
``the one shaped like an airplane''). They can also refer to the visual elements using associated
metadata such as rating or prime-eligibility (e.g., ``five-star one'',
``prime-eligible one'') or by their relative positions (e.g., ``middle one''). They can
also refer to visual entities from historical context. The dialogue agent can
automatically resolve the relevant visual entities from the context, determine
the next action, and fill the slots. For example, \emph{``Zoom on the red-striped shirt''} calls the \textit{ZOOM} action with the argument being the entity representing the ``red-striped shirt'' in the visual context. We enable multimodal understanding capabilities by introducing a new visual grounding model that can reason over the visual context given the user query and populate API arguments with visual entities\footnote{A video demo of our skill can be found at https://tinyurl.com/multimodal-conversations}.

Our main contributions are as follows:
\begin{enumerate}
\item We propose a novel multimodal conversational system that considers screen
context, in addition to dialogue context, while deciding the agent’s next
action.
\item The proposed visual grounding model takes both metadata and images as input
allowing it to reason over metadata and visual information. 
\item Our solution
encodes the user query and each visual entities
and then computes the similarity between them. To improve the visual entity encoding, we introduced query-guided attention and entity self-attention layers.
\item For training data, we create a multimodal dialogue
simulator. In addition, to capture real user behavior and collect more realistic data, we run
a large scale Mechanical Turk (MTurk) survey ($\sim$95K samples). 
\item Our model is able to achieve $\sim$85\% accuracy on difficult
crowd-sourced test set and we perform a thorough ablation study and show the contribution of each factor in model. 
\end{enumerate}

\section{Related Works}
In essence, our problem requires a combination of (i) a visual grounding solution that can exploit additional metadata to select an object based on a referring expression and (ii) a multimodal dialogue solution that can utilize dialogue and visual history to label the user intent and perform the implied tasks on a shared interactive medium. We review the related work in each of these topics separately.

{\bf Visual Grounding} Until recently, state-of-the-art visual grounding models
relied on combining the data from two modalities -visual and natural language- using various early fusion techniques \citep{yu2018mattnet,
anderson2018bottom}. Over the last couple of years, these methods have been
surpassed by transformer-based solutions that are pre-trained on various
combinations of image and text masking tasks \citep{su2019vl, lu2019vilbert,
li2020oscar, yu2020ernie, hu2021unit}. Our setting differs significantly from
the commonly studied setup due to (i) model having access to the exact location
of each object, which removes the need for object detection -a fundamental component of visual grounding- and (ii) additional metadata about the entities on the screen, which might contain both visual (e.g., color, material) and non-visual (e.g., prime eligibility, brand name) information about these entities. Thus, we cannot utilize other visual grounding solutions out of the box, and propose an attention based metadata injection mechanism that we present in Section 4.

{\bf Task-oriented Multimodal Dialogue Systems} Our visual grounding solution
enables selection of an item based on visual cues, but a multimodal shopping
experience involves tasks that require higher order reasoning, such as
co-referencing and disambiguation. Compared to the rich literature in
multimodality, the research multimodal dialogues is rather sparse. One line of
work considers visual question answering (VQA) tasks over a dialogue,
\citep{das2017visual, kottur2019clevr, de2017guesswhat, tu2021learning}. An
another line of work are the fruits of the SIMMC challenge
\citep{moon2020situated} which considers a multimodal conversational shopping
experience that involves an experience that is close to ours. In this line of
work, the literature relies on large scale transformers (e.g., GPT-2
\citep{radford2019language} and BART \citep{lewis2020bart}) that generates a
system action and a response based on the current and past utterances (i.e., the dialogue context) and the multimodal context \citep{kottur2021simmc}. Our approach involves building a visual understanding solution on top of an established conversational AI solution \citep{acharya-etal-2021-alexa} and thus we cannot rely on these generation based solutions.

\section{Data}
One of the major bottlenecks for building a multimodal goal oriented system is
lack of publicly available datasets. Although there are few, none of them fully
capture all variations that we want to support in our framework. For example,
the SIMMC challange~\citep{moon2020situated} dataset is not usable because (i)
out of all the APIs in this dataset, only one (\emph{FocusOnFurniture}) requires
visual context for correct prediction and that is referenced with an ordinal number most of the time. It does not contain examples with visual attributes like pattern or shape, except a few utterances for referencing based on color; (ii) DSTC9 API annotations only provide the dominant API even though there could have been multiple API calls between two turns. This issue, if not fixed, can lead to model confusion.

Due to the lack of readily available datasets, we explore two different
approaches for dataset creation. The first approach involves a dialogue
simulation technique with multimodal context that primarily uses the available
catalogues and associated metadata to create a domain-specific multimodal
conversational dataset. This approach enables creating seed data for model
training for any new domain without much effort. However, the models built may not be very robust. Hence, we complement our data with a second approach, where we design an MTurk collection pipeline to capture the actual user behavior while interacting with visual context. This allows us to collect a more realistic dataset with more language variations.

\subsection{Multimodal Dialogue Simulator}
To ensure a large degree of variation in the training data, we implement a
multimodal dialogue simulator. It can generate a large number of synthetic
conversations that capture numerous variations in how a visual entity is selected
(e.g., by name, position) and what operations are applied to it (e.g.,
selection, zoom-in). It also generates variations in the visual contexts and
examples of interactions spanning multiple turns (e.g., comparison, intent
carryover). The simulator works by randomly generating screen layouts with
visual entities and subsequently simulating user interactions with entities on the screen or with other entities mentioned recently. The method of operation is as follows:
\begin{enumerate}[leftmargin=*, noitemsep]
\item Randomly select up to two entities associated with components on the current simulated screen.
\item Randomly decide the method by which to refer to the chosen entities (e.g.,
color, position, item type). Generate a phrase that references the object (e.g., \emph{``the one on the right''}, \emph{``the blue one''}).
\item Check if the phrase uniquely identifies each of the entities among the ones visible on screen. If not, repeat step 2.
\item Randomly pick an action or a sequence of actions (e.g., in the case of comparison or intent carryover) to be performed on the entities. Generate a simulated sequence of utterances, for example \emph{``Show me the left one. Is it cheaper than the red one?''}.
\item Update the visual state based on the execution of the selected actions.
\item Repeat until MAX\_DIALOGUE\_LENGTH is reached.
\end{enumerate}

Simulated interactions include inquiring about properties of objects (e.g.,
\emph{``What is the material for the left one?''}), comparing objects, as well
as taking actions on them (e.g., \emph{``Add the red one to the cart''}). In
each interaction, objects can be referenced by metadata (e.g., name, price,
rating) or by their location on screen (e.g., \emph{``left one'', ``last one''}). Besides
one-turn requests, simulator also generates multi-turn conversational features
like anaphoric referencing (e.g., \emph{``What is the price of the rightmost
one? Add it to my cart''}) and intent-carryover (e.g., \emph{``What is the
rating of the blue one? How about the green one?'')} leading to a diverse
conversational dataset. Implementing the dialogue simulator also allows us to
quickly bootstrap a basic model without any additional training data. One limitation of simulator, however, is that
it can only generate data based on metadata attributes present in catalogue. As
a result, referencing based on shapes or patterns cannot be generated as these
attributes are typically not present in catalogue. We rely on MTurk collection to mitigate this problem, which we discuss next.


\label{sec:simulator}

\subsection{MTurk Collection}
\begin{figure}
\centering
\begin{subfigure}{.5\textwidth}
  \centering
  \includegraphics[width=.9\linewidth]{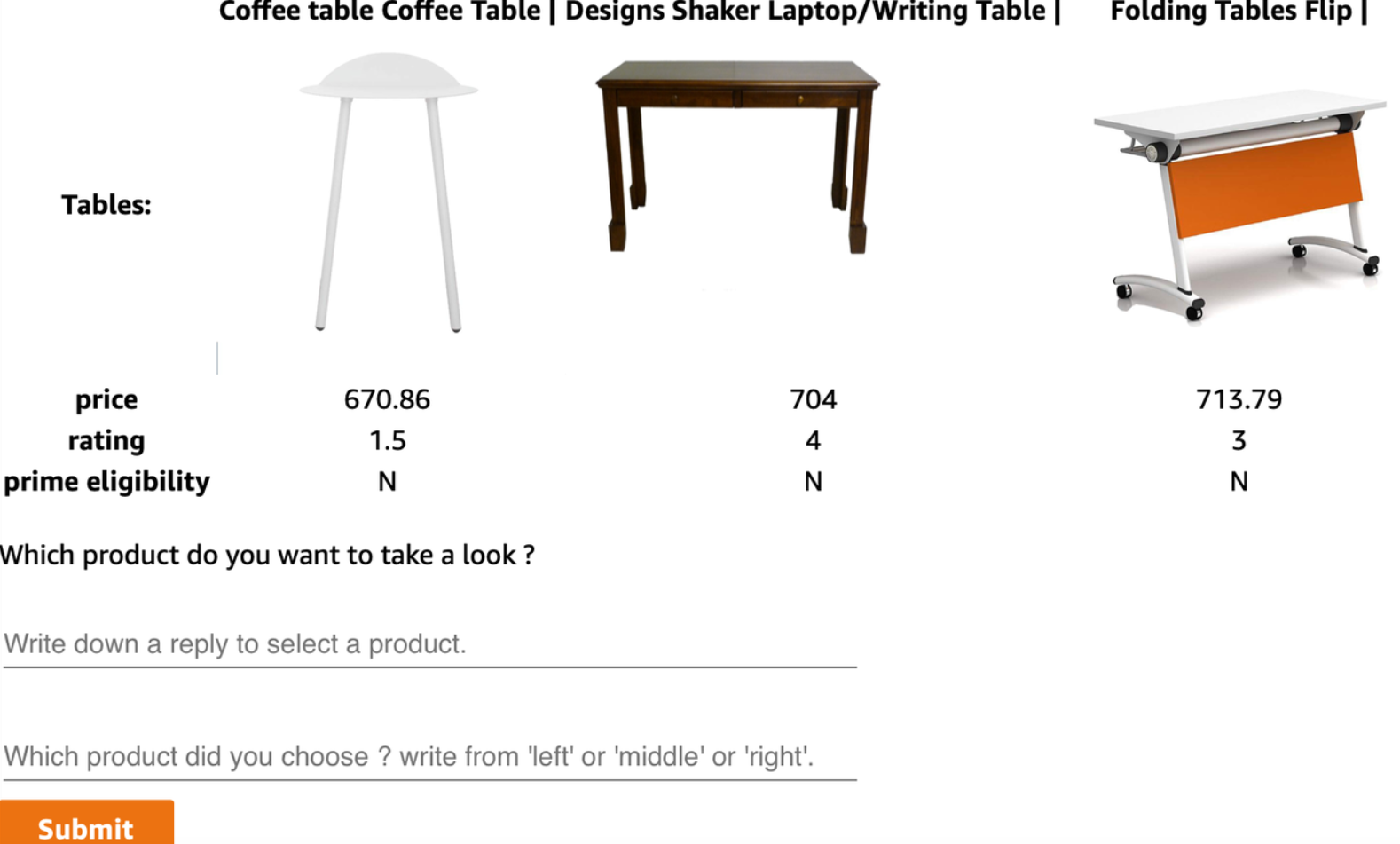}
  \caption{Amazon Mechanical Turker Interface}
  \label{fig:mturk1}
\end{subfigure}%
\begin{subfigure}{.5\textwidth}
  \centering
  \includegraphics[width=.9\linewidth]{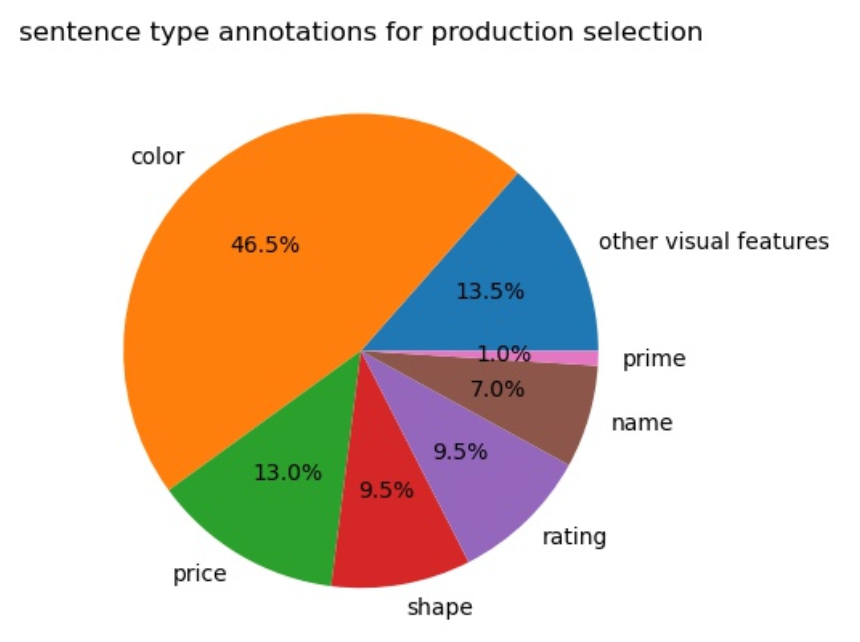}
  \caption{\small MTurk Data Distribution}
  \label{fig:mturk2}
\end{subfigure}
\caption{\small MTurk Multimodal Data Collection}
\label{fig:test}
\end{figure}

The variations covered in the simulated data may not be enough to build
very robust models. Moreover, training data should be as possibly close as it can be to real user behavior, and we are interested in exploring the variations of user utterances that refer to visual entities. Motivated by these factors, we crowdsourced data collection
using MTurk in a restricted setup to improve the robustness of our models and also obtain a more realistic evaluation test dataset.
We create an interface that
presents MTurkers three images representing three products along with their
name, price, rating, and prime eligibility information. We then ask the MTurkers to
provide an utterance to select a product among these three products as well as
the position of the referred product (left, middle or right). The web
interface is shown in Figure~\ref{fig:mturk1}.
We provide close to 20K different product combinations and for each combination we
assign five different MTurkers to complete the task, which results in
100K product selection utterances in total. In order to see what type of attributes
MTurkers usually use to select the product, we also manually annotate the utterance type among a small user-group with a subsample of 200 instances. We
divide the selection sentence types into selection by color, price, shape,
rating, name, prime eligibility, and miscellaneous visual features. The
annotation results are shown in Figure~\ref{fig:mturk2}. We find that 69.5\% utterances use visual features to select the particular product, indicating the need for multimodal models.
\label{sec:mturk}

\section{Approach}

\begin{figure}
\centering
\begin{minipage}{.5\textwidth}
  \centering
  \includegraphics[width=1.0\linewidth]{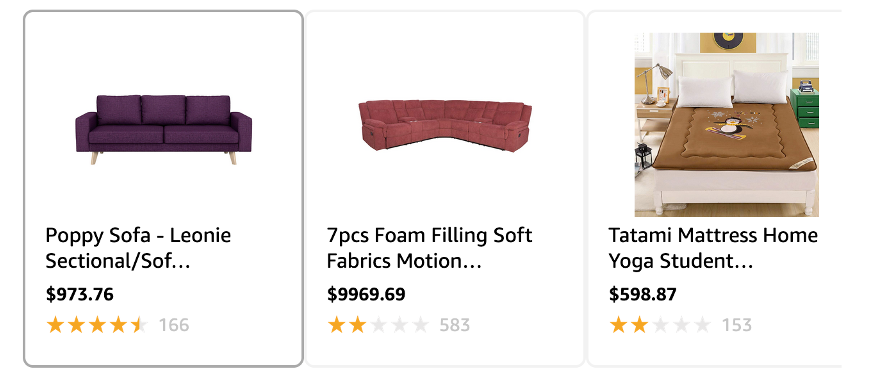}
  \captionof{figure}{\small Schema Based Screen Context}
  \label{fig:schema_vc}
\end{minipage}%
\begin{minipage}{.5\textwidth}
  \centering
  \includegraphics[width=0.5\linewidth]{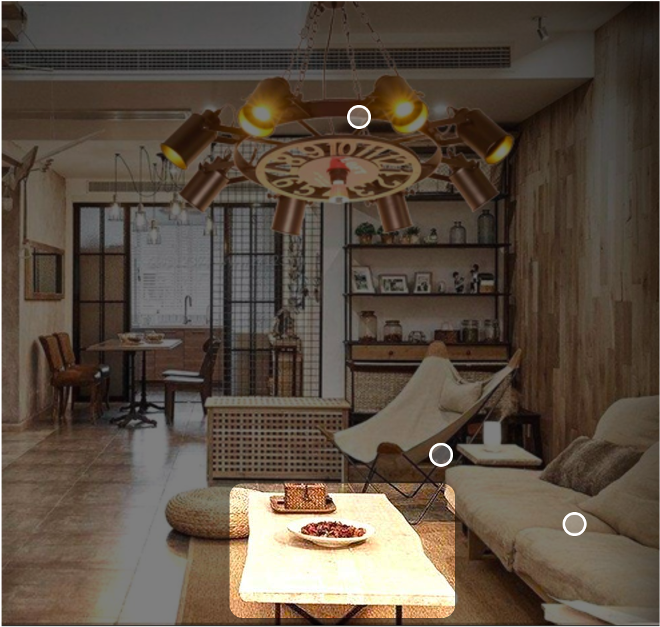}
  \captionof{figure}{\small Full Image Screen Context}
  \label{fig:image_vc}
\end{minipage}
\end{figure}

Our proposed model for grounding the user query with respect to visual context
consists of three main components: the query encoder, the visual entity encoder,
and the candidate scorer. At a high level; (i) we
first encode the query and each visual entity in a candidate set; (ii) next, we
compute a score for each (query, visual entity) pair; and (iii) finally we choose the
highest scoring entity. In the next few subsections, we 
define the notion of visual context and visual entity and describe how we create
the candidate set. We then discuss the details of each component of the model architecture. 

\subsection{Visual Context and Visual Entity}

In a goal-oriented dialogue system, the screen content is populated by the
back-end service providers after executing the API predicted by the agent. The
screen layout is typically defined using schemas. There are generally multiple
schemas used within an experience and each schema may contain multiple visual
elements. For instance, for a shopping experience, the search page may contain
three sofa images, a next page button and a go back button for a total of 5
visual elements. We consider each such visual element as one visual entity.
Visual context corresponding to a particular turn is a collection of all the
visual entities within a page. During a conversation, the user will typically
interact with the screen (verbally or through touch) via these visual elements,
and hence those are likely candidates for visual grounding.
Figure~\ref{fig:schema_vc} shows an example of a schema-based screen, with 3
sofa entities. Note that a visual entity may have not only the visual modality
(i.e., image) but also textual modality (e.g., metadata such as name and price) as seen in
the Figure.

However, often the schema may not be as well defined, or the full screen may
contain just an image. In these cases, we use an off-the-shelf object
detector to annotate the contents on the screen. We consider
each detected object as a visual entity and the set of all the detected objects
as the visual context. This is necessary as during the conversation, a user may
be interested in items embedded within the image on the screen. For instance, in
Figure~\ref{fig:image_vc} the user may ask the agent to zoom on the coffee
table. We utilize an Amazon internal tool called \href{https://www.amazon.com/stylesnap}{StyleSnap} for object detection. It provides the bounding box, a category label and a finer-label for each of the detected objects. As before, these visual entities also have both visual modality and the corresponding textual modality.

To construct the candidate set, we include all the visual entities from the
current screen. Furthermore, we also include historical visual entities from
previous screens in earlier turns of the conversation. This allows the user to
refer to past entities, for example, \emph{``Compare the red sofa with the grey
one that you showed earlier''}. We apply de-duplication on the candidate set.

\subsection{Model Architecture}

\begin{figure}
\includegraphics[width=1.0\linewidth]{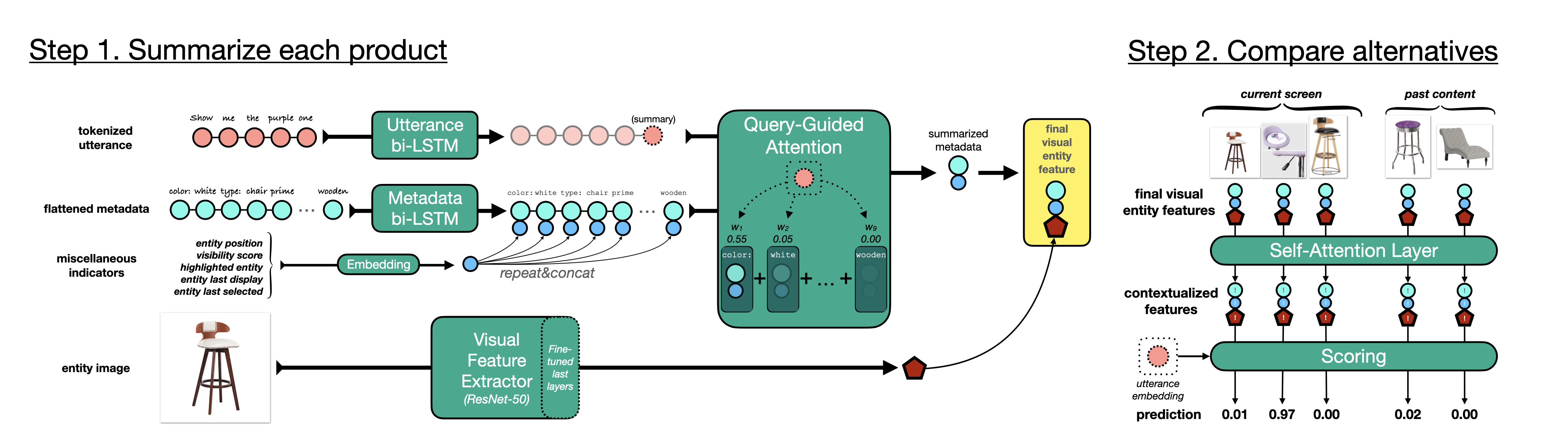}
\caption{\small Model Architecture}
\label{fig:model}
\end{figure}

We present the full architecture in Figure~\ref{fig:model}. We describe its
various components in the following subsections.

\subsubsection{Query Encoder}

The query consists of multiple components: dialogue context, the
current user utterance, and the API argument that needs to be filled. An API can
have multiple arguments; the visual grounding model needs to uniquely point to
the referred entity for each argument. The dialogue context is flattened and
encoded using a bi-directional LSTM. The current user utterance is also encoded with a (separate) bi-directional LSTM. For the API argument, we encode the argument name and argument type using dense embeddings. The final query representation is obtained by concatenating the final hidden states from dialogue context and current utterance LSTMs together with the API argument embeddings.

\subsubsection{Visual Entity Encoder}

A visual entity has two main components: the metadata and the image. Metadata
features are required as a user can refer to visual entities via information
present in metadata such as name, rating, and prime eligibility. For instance, users can
say
\emph{``What are the dimensions of the
four star one''}. Image features are also required as the
users can also refer to entities via visual characteristics. For instance, users
can say
\emph{``Add the striped chair to my
cart''}. In addition, we also include the spatial properties of the entity as
displayed on screen. This includes the location and visibility of the visual
entity on screen. This allows users to refer visual entities based on their
geometry in the visual context, for instance,
\emph{``Tell me details about the third one''}. Visibility is needed so that the
model can ignore partially visible entities (that might have appeared during
touch based interactions with the screen). This allows the model to learn an
optimal acceptance threshold for visibility from the data, without having to
hard-code a fixed value. We also encode the state of the visual entity where
state is a binary feature indicating whether an entity is highlighted or not on
the screen. An entity can be highlighted when the user or the agent mentions it.
This allows the model to resolve anaphoric references to that entity, for
instance, \emph{``Add it to my cart''}. Finally, we also include memory features
indicating the turn when the entity was last visible and when it was last
selected by the user. This allows the model to distinguish past entities from
current screen entities and to bias towards recently-referenced entities.

We encode the metadata using a bi-directional LSTM over a flattened sequence of
metadata property names and their values. The flattened metadata sequence can be
arbitrarily long,
since there could be large number of
attributes. To allow the model to focus on the desired attributes, we introduce
query-guided attention. 
The encoded query attends to each token in
the metadata sequence, and we compute the query-attended metadata encoding. For the
image, we extract visual features using the learned top of the
pre-trained ResNet-50 model. For encoding the entity location, we extract the
position of the entity on screen and encode it using sinusoidal positional
embeddings~\cite{vaswani2017attention}. We assume a linear 
layout of
visual entities thus only encode the $x$-coordinate positions.
In general, the 4-D bounding box can directly be encoded using sinusoidal
embeddings similar to \cite{hu2018relation}. For visibility, we bucketize the
visibility score and embed it using 1-hot vector. For encoding the highlighted
state, we use a small dense embedding. Finally, memory features are encoded with
sinusoidal positional embeddings as well. All the feature encodings are
concatenated to form the joint representation of the visual entity. In order to
make each visual entity aware of all the other entities in the visual context,
we add a self-attention layer on top of the joint representation. This allows
the model to refer entities based on their relative positions on the screen, for
instance, \emph{``Select the rightmost one''}. This also allows the model to
perform comparative reasoning, for instance, 
 \emph{``Is the highest rated one prime eligible''}. The self-attended encoding
is the final representation for visual entities.

\subsubsection{Candidate Scorer}

The scorer is a simple bi-linear attention layer. We compute the attention
between the query encoding and visual entity encoding for each visual entity in
the candidate set. We use resulting attention scores as relevance scores and the entity with the highest score is returned as the output.

\section{Experiments and Discussion}
We present the results of our experiments on our furniture shopping dataset that contains simulated data, as well as data from MTurkers. The product catalog for our dataset is built using
a small subset of Amazon catalogue with close to 50K furnitures. We generate
the simulated dataset using the multimodal dialogue simulator described in
Section~\ref{sec:simulator}. The simulated training and test set contain about
300K and 19.5K examples respectively. We use non-overlapping subsets
of catalogues for training and test data generation. To improve diversity in
training set and collect a more realistic test set, we augment the simulated
data with an MTurk collection following the setup described in
Section~\ref{sec:mturk}. We collect about 95K examples, which we split into $8:1:1$ 
for training, development, and test without overlap.
We use FastText for word embeddings~\cite{mikolov2018advances}. LSTM hidden size is 50 for query and metadata encoders.
For image encoder, we use a ResNet-50 model adding a 50-dim top layer and fine-tuning the top-3 layers. For encoding positions and highlighted state, we use embeddings of size 50 and 5 respectively. The visibility score is separated into 20 buckets. We train the models using Adam optimizer without weight decay and minimize the cross-entropy loss. We use a batch size of $128$ and a learning rate of $0.001$.


\renewcommand{\arraystretch}{1.5}
\begin{table}[]
\centering
\scriptsize
\begin{tabular}{c|c|c}
\toprule[0.5mm]
\multirow{2}{*}{\textbf{Model}} & \multicolumn{2}{c}{\textbf{Accuracy (\%)}}   \\ \cline{2-3}
&  Simulated Test Set & MTurk Test Set \\ \hline \hline
Random Baseline & 12.68  & 33.44 \\ \hline
\makecell{Proposed Model \\ (Only current screen context)} & 84.89   & 84.23  \\ \hline
\makecell{Proposed Model \\(Screen context from last 3 turns)} & 97.73 & 85.13  \\
\bottomrule[0.5mm]
\end{tabular}
\vspace{3mm}
\caption{\small 
MTurk set only contains examples for referencing visual entities from current
screen. Simulated set has examples for user reference both current screen and
past screens.}
\label{table:results}
\vspace{-3mm}
\end{table}

Table~\ref{table:results} shows the performance of the proposed visual grounding
model on simulated and MTurk test sets. We train two variants of the proposed
model, one where the candidate set is created only from the current screen and
the other where it also includes past visual entities. The model architecture is
same in both the cases. For baseline, we randomly choose a visual entity from
candidate set. Note that, the random baseline performance is better on the MTurk
set. This is because the way the MTurk collection is set up, it only contains
examples with current screen context and thereby results in smaller candidate
set. For the same reason, the performance on MTurk test set is largely unchanged
whether or not we train the proposed model with historical visual context.
However, on the simulated set, including historical entities gives significant
improvement, primarily for cases where the user tries to reference a previously
shown entity, e.g., \emph{``Is the blue one cheaper than the previous red
one''}. Our proposed model is able to achieve reasonable accuracy of $\sim$85\% on a realistic and relatively difficult MTurk set.

We also conduct ablation studies across different dimensions to gauge impact
of each component on model performance. Table~\ref{table:ablation} shows the
results of our ablation study. We observe that:
\begin{enumerate}[leftmargin=*, noitemsep]
  \item The MTurk data and simulated data are complimentary.
Training with only simulated data leads to poor performance on MTurk set
and vice-versa. More importantly, when the model is trained with combined data,
it leads to knowledge transfer resulting in better performance on both test
sets.

  \item Both metadata and visual features contribute to the
model accuracy. When trained without metadata features, the performance drops
significantly on both test sets. The resulting model cannot resolve entities
based on metadata attributes such as name and price. However, it can still
reference based on visual features like color, which is reflected in the almost
unchanged accuracy for reference by color. On the other hand, when trained
without visual features, color reference accuracy drops significantly. With this
model, the drop in overall accuracy is much higher on MTurk set compared to the
simulated set. This is because $\sim$70\% examples in the MTurk set require
referencing via visual characteristics while the simulated set has only
$\sim$17\% such cases.

  \item 
We see a $\sim$4\% accuracy improvement from the query guided attention. It helps
the model to focus on the right attributes in the metadata based on the user query.
Entity self-attention layer makes entities aware of each other and
gives another significant performance boost. We see higher gains on the MTurk
set as it has many examples requiring comparative reasoning, e.g., \emph{``Show
me the highest rated one''}. Changing the loss function from binary
cross-entropy to cross-entropy allows the model to better discriminate entities
in the candidate set and provides an another 1--2\% improvement.
\end{enumerate}

\renewcommand{\arraystretch}{1.5}
\begin{table}[]
\centering
\scriptsize
\begin{tabular}{c|c|c|c}
\toprule[0.5mm]
\multirow{2}{*}{\textbf{Model}} & \multicolumn{3}{c}{\textbf{Accuracy (\%)}}   \\ \cline{2-4}
&  \multicolumn{2}{c|}{Simulated Test Set} & MTurk Test Set \\ \hline
&  Overall & Referencing by Color & Overall \\ \hline \hline
Random Baseline & 33.52 & 27.23 & 33.44 \\
\toprule[0.3mm]
\multicolumn{4}{c}{\textbf{Data Source}} \\
\bottomrule[0.3mm]
Simulated + MTurk data & 96.22 & 84.68 & 85.89 \\ \hline
$-$ MTurk data & 95.58 & 82.71 & 57.33 \\ \hline
$-$ Simulated data & 43.00 & 55.32 & 82.50 \\
\toprule[0.3mm]
\multicolumn{4}{c}{\textbf{Feature Type}} \\
\bottomrule[0.3mm]
Metadata + Visual features & 96.22 & 84.68 & 85.89 \\ \hline
$-$ Metadata features & 80.66 & 83.61 & 62.43 \\ \hline
$-$ Visual features & 91.47 & 54.83 & 72.63 \\
\toprule[0.3mm]
\multicolumn{4}{c}{\textbf{Model Component (Without Visual Features)}} \\
\bottomrule[0.3mm]
Vanilla concatenation & 82.00 & 49.63  & 52.42 \\ \hline
$+$ Query-Guided attention & 86.27  & 50.12 & 56.55  \\ \hline
$+$ Entity self-attention & 90.59 & 51.47 & 70.53 \\ \hline
$+$ Cross-Entropy loss & 91.47  & 54.83  & 72.63 \\
\bottomrule[0.5mm]
\end{tabular}
\vspace{3mm}
\caption{\small Ablation Experiments}
\label{table:ablation}
\vspace{-3mm}
\end{table}

\section{Conclusion}
We introduce a model that can resolve visual entities within
dialogues. The new model can perform visual reasoning by comparing various entities and
choosing the correct visual entity based on dialogue and visual context. To
train and test model, we introduce a new multimodal dialogue simulator as
well as a mechanism for crowd-sourcing multimodal data. On data that are close
to real world, that relies on actual products from Amazon.com and utterances collected
through crowd-sourcing, we show that our models have high accuracy of 85\%. We
also provide a detailed ablation study to show how each of component contribute to
this accuracy. In
the future, we plan to integrate visual context and dialogue context into a
single model rather than having separate representation as presented in this
work.



\bibliography{mm}
\bibliographystyle{plain}

\end{document}